\documentclass{article}
\pdfoutput=1
\usepackage{ijcai22}

\usepackage{times}
\usepackage{soul}
\usepackage{url}
\usepackage[utf8]{inputenc}
\usepackage[small]{caption}
\usepackage{graphicx}
\usepackage{amsmath}
\usepackage{amsthm}
\usepackage{booktabs}
\usepackage{algorithm}
\usepackage{algorithmic}
\usepackage{multicol}
\usepackage{bm}
\usepackage{dsfont}
\usepackage{float}
\usepackage{amssymb}
\usepackage{bbding}
\usepackage{multirow}
\usepackage[perpage]{footmisc}
\usepackage[colorlinks=true,
            linkcolor=bred,
            urlcolor=citecolor,
            citecolor=citecolor,
            anchorcolor=magenta]{hyperref}
\usepackage{siunitx}
\usepackage{paralist}
\usepackage{listings}
\usepackage{pdfpages}

\usepackage{multirow}
\usepackage{colortbl}
\usepackage{xcolor}

\usepackage{subfigure}
\newtheorem{definition}{Definition}

\newcommand{\appendixhead}%
{\centering\textbf{\huge Appendices}
\vspace{0.25in}}

\def\ba{{\mathbf{a}}}
\def\bo{{\mathbf{o}}}
\def\bs{{\mathbf{s}}}
\def\bv{{\mathbf{v}}}
\def\bz{{\mathbf{z}}}
\def\bx{{\mathbf{x}}}
\def\by{{\mathbf{y}}}

\definecolor{citecolor}{HTML}{0071bc}
\definecolor{ogreen}{HTML}{2E7D32}
\definecolor{bred}{HTML}{BF360C}
\definecolor{newbrown}{HTML}{795548}

\hypersetup{breaklinks=true,colorlinks=citecolor}

\title{Accelerating Representation Learning with View-Consistent Dynamics in Data-Efficient Reinforcement Learning }

\author{
Tao Huang$^1$\footnotemark[1]
\and
Jiachen Wang$^1$\footnotemark[1]\and
Xiao Chen$^{2}$
\affiliations
$^1$ShanghaiTech University\\
$^2$Zhejiang University\\
\emails
\{huangtao1, wangjc2\}@shanghaitech.edu.cn,
xiaochen.it@gmail.com
}

\begin{document}

\maketitle

\renewcommand{\thefootnote}{\fnsymbol{footnote}}
\footnotetext[1]{Equal contribution.} 
\renewcommand{\thefootnote}{\arabic{footnote}}

\begin{abstract}
  Learning informative representations from image-based observations is of  fundamental concern in deep Reinforcement Learning (RL). 
  However, data-inefficiency remains a significant barrier to this objective. To overcome this obstacle, 
  we propose to accelerate state representation learning by enforcing view-consistency on the dynamics. Firstly, we introduce a formalism of Multi-view Markov Decision Process (MMDP) that incorporates multiple views of the state into traditional MDP. Following the structure of MMDP, our method, \textbf{V}iew-\textbf{C}onsistent \textbf{D}ynamics (\textbf{VCD}), learns state representations by training a view-consistent dynamics model in the latent space, where views are generated by applying data augmentation to states. Empirical evaluation on DeepMind Control Suite and Atari-$100$k demonstrates VCD to be the SoTA data efficient RL algorithm on visual control tasks.
\end{abstract}

\section{Introduction}
\label{intro}
Deep Reinforcement Learning (RL) harnesses the expressive power of deep neural networks and the long-term reasoning ability of RL to solve sequential decision-making problems~\cite{Mnih2015HumanlevelCT}. Recent years have witnessed the sensational progress of it in various complex control tasks, such as playing video games~\cite{hafner2019dream}, robotic control~\cite{kalashnikov2018scalable} and autonomous driving~\cite{ShalevShwartz2016SafeMR}.

Despite the notable success of deep RL, recent studies have revealed that data-/sample- inefficiency severely impedes its performance when learning from high dimensional observations~\cite{Lake2016BuildingMT}. This remains a significant barrier to the real-world applicability of deep RL, where collecting experiences is often costly and time-consuming~\cite{dulac2019challenges}. For instance, a successful RL agent requires several months to develop a decent grasping skill, standing sharply in contrast to the human-level efficiency~\cite{kalashnikov2018scalable}. Accordingly, elevating data efficiency is of paramount importance for the broader progress of deep RL.

Many existing works approach this goal by augmenting deep RL with self-supervised tasks. The motivation of that is two-fold: (i) the potentially sparse reward signals are incapable of learning a good state representation with limited data~\cite{kostrikov2020image,laskin2020curl}; (ii) Self-Supervised Learning (SSL) unleashes the potential of massive unsupervised signals for representation learning, which has achieved remarkable performance in downstream vision and language tasks, particularly in low data regimes~\cite{chen2020simple,grill2020bootstrap}. Beyond that, there are proliferative paradigms of designing SSL tasks in RL due to its interactive and temporal-correlated training mechanism, such as maximally preserving predictive information~\cite{oord2018representation,lee2020predictive}, modeling dynamics~\cite{Jaderberg2017ReinforcementLW,schwarzer2020data,yu2021playvirtual} and discriminating features of instances at the spatial or temporal level~\cite{laskin2020curl,stooke2021decoupling}.

In this work, we propose to further accelerate representation learning in RL by enforcing view-consistency on the learnt dynamics model within a latent space. 
We posit that learning a view-consistent dynamics can encode transition-relevant information into the state representations, thus enabling an RL agent to efficiently exploit the environment dynamics into its decision-making process.
We formalize the above idea with a new decision-making framework called Multi-view Markov decision process (MMDP), which takes the view space of the state into consideration. With some realistic assumptions, we point out that the transition dynamics holds the same dynamics characteristics over multiple views given a state. 

\begin{figure*}
\begin{center}%
{\includegraphics[scale=0.7]{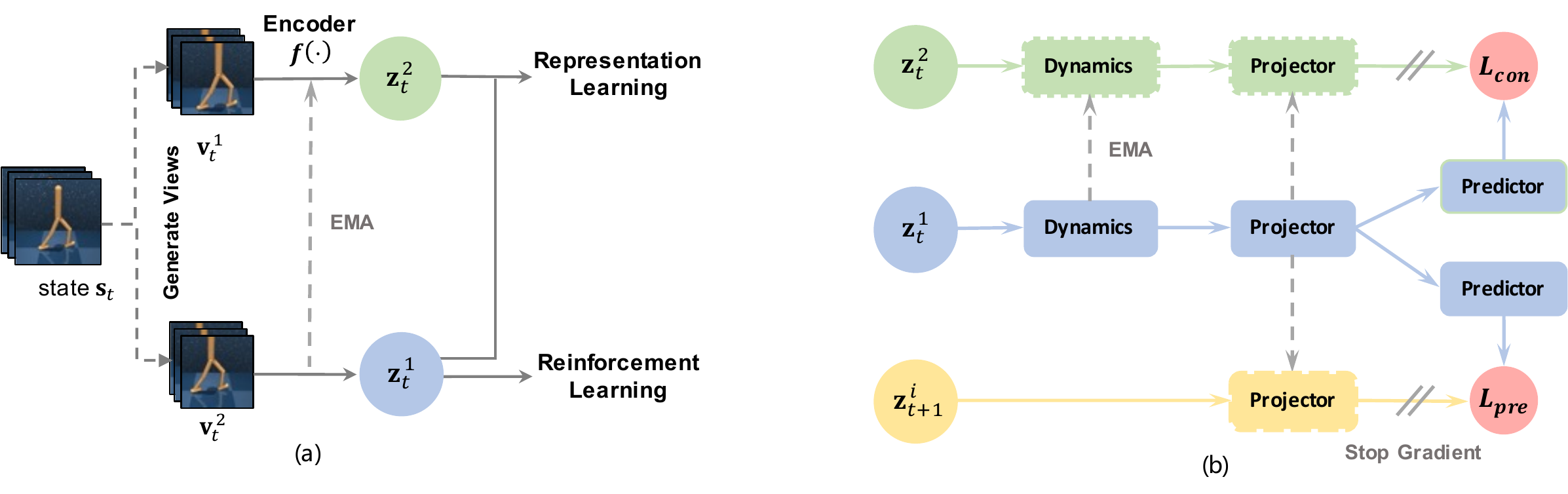}}
\vspace{-5mm}
\end{center}
\caption{RL policy learning with VCD: \textbf{(a) Overview of our framework.} Different views are generated from a given underlying state and encoded to view representations. Both views are utilized in our auxiliary task for representation learning. \textbf{(b) Illustration of our auxiliary objective.} The view-consistency objective is calculated between the representations of two views predicted by the dynamics module. And the prediction objective is realized by enforcing the predicted next-state representations towards the ground-truth state representation.}
\label{fig:framework}
\end{figure*}

Following the structure of MMDP, our method, View-Consistent Dynamics (VCD), learns state representations by training a view-consistent dynamics model within a latent space, where views are generated by applying data augmentation to the same underlying states. Composing it with RL objective teases out the final objective in the whole course of policy learning. We demonstrate our framework in \figurename{ \ref{fig:framework}}.

We evaluate VCD on a series of pixel-based control tasks from the DeepMind control suite~\cite{tassa2018deepmind} to Atari games~\cite{ale}.
The empirical evaluation shows that our VCD agents outperform prior state-of-the-art baselines on different evaluation metrics. We also conduct extensive ablation studies to validate the efficacy of the view-consistent dynamics. 

We highlight our main contributions below:
\begin{itemize}
    \item We propose a new framework MMDP that extends a traditional MDP state to it multiple views and present a novel but simple method VCD that exploits the structure of MMDP to accelerate the representation learning in RL . 
    
    \item We demonstrate that VCD agents outperform prior state-of-the-art baselines on the widely used pixel-based control tasks from DMControl and Atari games in terms of both data-efficiency and asymptotic performance. We also adopt the recently proposed rigorous deep RL evaluation metrics as our benchmarking methodology. 
    
    \item   Through careful ablation studies, we verify the efficacy of the view-consistency module itself and other 
    implementation details in incorporating SSL to deep RL.
\end{itemize}

\section{Related Work}\label{sec:related work}  
Many algorithms have been proposed to improve the data efficiency of deep RL agents that take input as high dimensional observations like images. We classify the existing methods into three categories as follows. 

\paragraph{Build world models.} The first line of work explicitly builds world models of the environment. Representative works include PlaNet~\cite{hafner2019learning}, Dreamer~\cite{hafner2019dream} and SLAC~\cite{lee2020stochastic} that perform planning or rollout-sampling in the latent space through a learnt world model. On the contrary, the dynamics model in UNREAL \cite{Jaderberg2017ReinforcementLW} are learnt to obtain compact state representations without planning, which shares a similar idea with our method. The main difference here is that our dynamics model is located in the representation space, avoiding the inefficiency incurred by pixel-level reconstruction.
\paragraph{Apply data augmentation.} The second line of works devotes to unleashing the potentials of data augmentation. For instance, CURL~\cite{laskin2020curl} learns contrastive unsupervised representations from visual observations, achieving high data-efficiency in DeepMind Control Suite and Atari games. The core of CURL is to generate key-query pairs through data augmentation for a contrastive loss. This idea is further explored in ATC~\cite{stooke2021decoupling} and ST-DIM~\cite{anand2019unsupervised} where a temporal contrast is adopted instead. We also perform an ablation study to the contrastive loss in our method in Section~\ref{subsec:ablation}. Beyond that, \cite{laskin2020reinforcement} observes that simply applying data augmentation on the input observations can greatly improve the data-efficiency. Their method, named RAD, is further extended by DrQ~\cite{kostrikov2020image}. It regularizes model-free RL algorithms with multiple augmented states (views). However, unlike DrQ, we use multiple views to regularize representation learning instead of RL. 

\paragraph{Design auxiliary tasks.} The third line of works leverage the recent advances in unsupervised and self-supervised representation learning~\cite{chen2020simple,grill2020bootstrap} by designing auxiliary tasks along with RL. One designing paradigm focuses on reconstruction-based objectives, such as future prediction in UNREAL and image-reconstruction in SAC-AE~\cite{yarats2021improving}. Another vine of works learns representations by discriminating features of instances at spatial or temporal level, including CURL and ATC. Beyond that, some works propose to harness the Markovian structure of the environment. For instance, CPC~\cite{oord2018representation} and PI-SAC~\cite{lee2020predictive} maximally preserves the predictive information in the state representations; SPR~\cite{schwarzer2020data} and PlayVirtual~\cite{yu2021playvirtual} predict future state representation by learning a latent dynamics model. {Our method bears some resemblance to SPR in learning dynamics, except that a new property, view-consistency, is being enforced on our model.}

\section{Method}\label{sec:method}
In this section, we propose our method named VCD to improve the data-efficiency in (pixel) RL. Our key idea is to accelerate the state representation learning by training a \emph{view-consistent} dynamics model in the latent space. To achieve this, we incorporate a self-supervised task into RL that includes two orthogonal but complementary parts: (i) train a latent dynamics model to predict the future view representations; (ii) force the dynamics model to be view-consistent. 

We first formalize Multi-view Markov Decision Process (MMDP) and the definition of $\emph{view-consistency}$ in Section~\ref{subsec: mmdp}. In Section~\ref{subsec:vc=dynamics}, we then introduce the \emph{view-consistent} dynamics model and show how it helps representation learning in RL. Finally, we discuss the practical concerns of implementing VCD in Section~\ref{subsec: impl-detail}.

\subsection{Multi-view Markov Decision Process}
\label{subsec: mmdp}
A Markov decision process (MDP) in RL is defined by a tuple $\langle \mathcal{S},\mathcal{A},\mathcal{P},r,\mathcal{\gamma}\rangle$, where $\mathcal{S}$ is the state space, $\mathcal{A}$ is the action space, $\mathcal{P}$ is the transition dynamics, $r$ represents the reward function whose element $r(\bs_t,\ba_t)$ is the reward collected by taking action $\ba_t$ at state $\bs_t$, $\gamma\in [0,1)$ denotes the discount factor. Crucially, we stack $l$ consecutive image-based observations $(\bo_{t-l+1},...,\bo_{t})\in \mathcal{O}^l$ as the fully-observed state $\bs_t$.
The agent's objective is to find a policy $\pi(\cdot|\bs_t)$ that maximizes the cumulative discounted return $\mathbb{E}_{\pi}\left[\sum_{t=0}^\infty\gamma^tr(\bs_t,\ba_t) \right]$. 

\begin{figure}
\begin{center}
\centerline{\includegraphics[width=8cm]{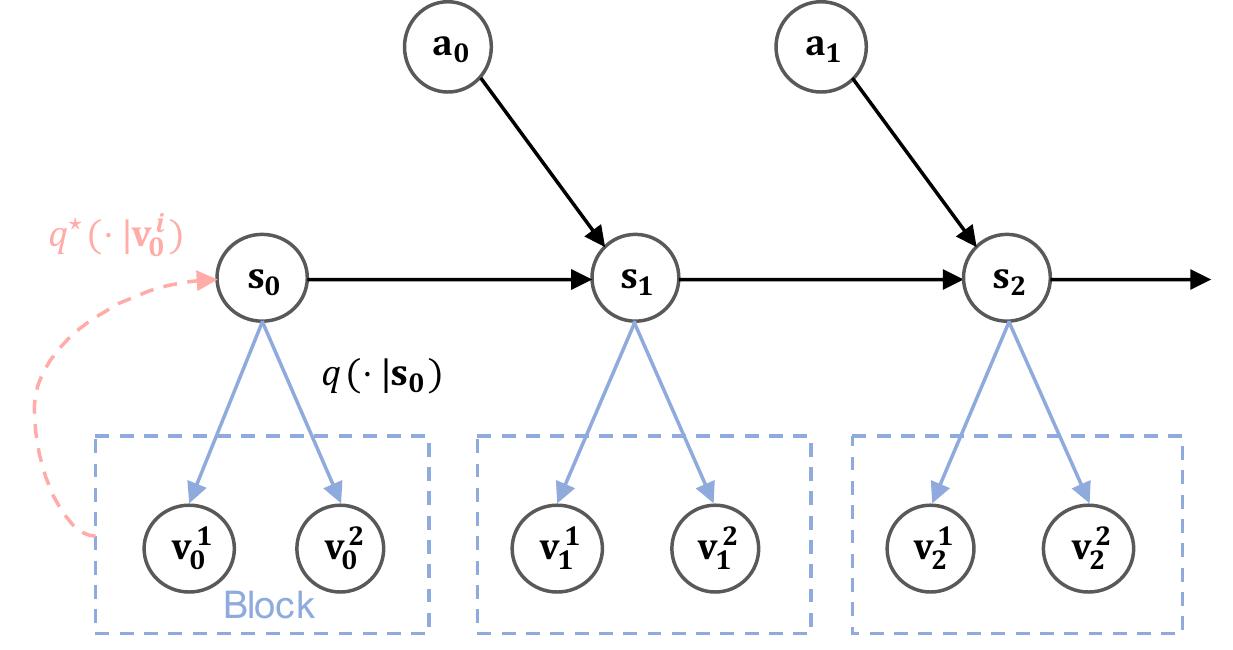}}
\caption{\textbf{The graphical model of MMDP}. The state $\bs_t$ generates views $\bv_t^i$ through the rendering function (blue arrow) $q(\cdot|\bs_t,\upsilon_t^i)$. The views $\bv_t^i$ are in a block that can be decoded back to into their underlying state $\bs_t$ through decoding function (red arrow) $q^\star(\cdot|\bv_0^i)$.}
\label{fig:mmdp}
\end{center}
\vskip -0.2in
\end{figure}

Data augmentation is wildly adopted in RL to improve data-efficiency .
These methods share a key insight that data augmentation unleashes the potential of the massive pixel observations (states). To further leverage its power, we propose to model the \textit{view space} $\mathcal{V}$ of the state. 
Specifically, the  views $\bv_t^{i}$ are emitted from one underlying state $\bs_t$ with the rendering function $q(\cdot|\bs_t)$. And data augmentation like rotation, crop or translation are all specific instances of such rendering functions.
We further assume a block structure~\cite{du2019provably} over the view space $\mathcal{V}$. That is, $\mathcal{V}$ can be partitioned into disjoint blocks $\mathcal{V}_s$, each containing the support of the conditional distribution $q(\cdot|\bs)$ and a perfect decoding function $q^\star:\mathcal{V}\rightarrow\mathcal{S}$ that maps $\bv_t^i$ to $\bs_t$. This is a realistic assumption. For instance, one can rotate an image to generate multiple views, and these views can be rotated back to restore the original images. We illustrate the ordinary MDP and the resulting \textbf{Multi-view MDP} (MMDP) in Figure~\ref{fig:mmdp}.

The distinct structure of MMDP implies a core property that guides our representation learning method: the transition dynamics is invariant to a state $\bs_t$ and its views $\bv^i_t$:
$$\mathcal{P}(\bs_{t+1}|\bs_t,\ba_t)=\mathcal{P}(\bs_{t+1}|\bv^i_t,\ba_t), \,i\in\{1,2\},$$
and we call it \emph{View-Consistency} of the transition dynamics $\mathcal{P}$. In other words, the block structure on the view space retains all the dynamics information among different views. In the following section, we extend this property in the latent space to boost state representation learning.

\subsection{How View-consistency Helps RL}
\label{subsec:vc=dynamics}

We first describe the latent dynamics model. Then we discuss how to exert view-consistent to this dyanmics model and how it helps representation learning in RL. 
\paragraph{Latent dynamics model.} Encoding predictive information into the state representation $\bz^s_t\in\mathcal{Z}$ aids policy learning in RL~\cite{oord2018representation}, where $\bz^s_t=f(\bs_t)$ is encoded by a feature encoder $f(\cdot)$. Some works realize this by training a latent dyanmics model (DM) $h(\cdot, \cdot)$ that predicts the transition dynamics $(\bz^s_t, \ba_t)\rightarrow\bz^s_{t+1}$ in the representation space $\mathcal{Z}$. Following structure of MMDP, we consider the DM that takes as input view representation $\bz_t^i=f(\bv^i_t)$ and an action $\ba_t$, which then predicts the future view representation $\bz_{t+1}=f(\bv_{t+1})$.\footnote{We remove the 
superscript here for clarity, as $\bz_{t+1}$ can denote any view representations given ground state $\bs_{t+1}$.} 

\paragraph{View-consistent dynamics model.} In our multi-view setting, the transition dynamics holds the view-consistency property in the state space. It is natural to ask:
\begin{center}
    \emph{What if a latent dynamics model holds view-consistency?}
\end{center}
To investigate this problem, we first provide a formal definition of the view-consistent DM as follows:
\begin{definition}[View-consistent dynamics model]
Given any two view representations $\bz_t^1$ and $\bz_t^2$ and action $\ba_t$, a latent dynamics model $h$ is view-consistent, if  for any time step $t$,
\begin{equation}\label{eq:distance}
\begin{gathered}
d\big(h(\bz_t^1,\ba),h(\bz_t^2,\ba)\big)=0,
\end{gathered}
\end{equation}
where $d$ is a distance metric in the representation space.
\end{definition}

Essentially, a view-consistent DM holds the same dynamics characteristics over view representations given a state. By enforcing view-consistency on the DM, the RL agent learns an encoder that extracts transition-relevant information. We posit that learning such a DM will accelerate representation learning in RL. To this end, we will restrict the DM with view-consistency. Designing details are discussed in the following section.

\begin{figure*}
\begin{center}
\resizebox{\textwidth}{!}{\includegraphics[width=1\hsize]{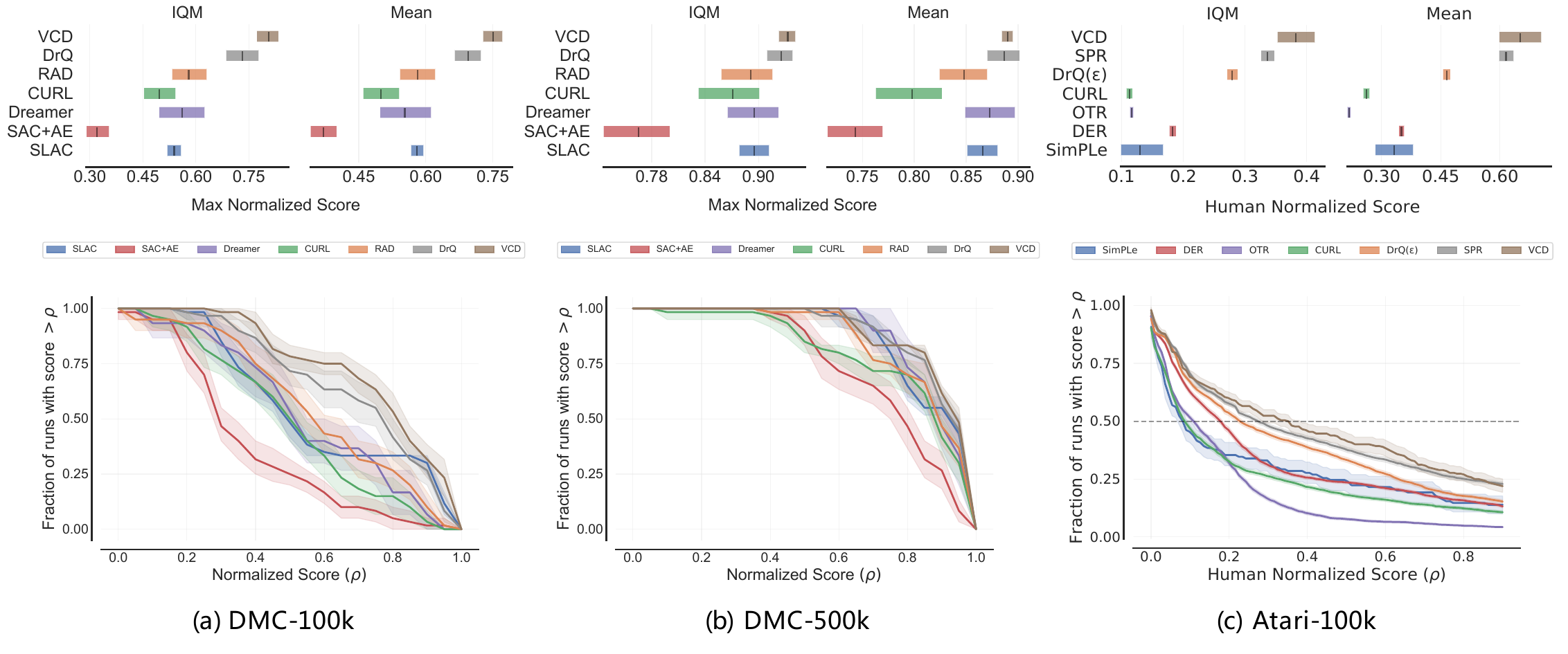}}
\vspace{-8mm}
\end{center}
\caption{Performance evaluation on DMC-100k, DMC-500k and Atari-100k. We run each task with 10 different random seeds. \textbf{First line. Aggregate metrics}, including Mean and Interquartile-mean (IQM) score of VCD and other state-of-the-art baselines with $95\%$ Confidence Intervals (CIs). The CIs are estimated using the percentile bootstrap with stratified sampling. \textbf{Second Line. Performance Profile} reflects the score-distribution across all runs, which is also more robust to outliers. Shaded regions show pointwise $95\%$ condifence bands based on percentile bootstrap with stratified sampling. The area under the performance profile corresponds to the mean.  }
\label{fig:main_all}
\end{figure*}

\subsection{Implementation Details}
\label{subsec: impl-detail}

\paragraph{Online and target architecture.} We consider a one-step transition $\left((\bv_t^1,\bv_t^2),\ba_t,r_t,\bv_{t+1}\right)$ in an MMDP. Following~\cite{grill2020bootstrap,schwarzer2020data}, we adopt two lines of nerworks: (i) online networks that include online encoder $f_{o}$, online DM $h_o$ and online projector $g_o$ parameterized by  $\theta_o$; and (ii) their counterpart in the target line parameterized by $\theta_m$.
The online (target) encoder maps view $\bv^1_t$ ($\bv^2_t$) into view representation $\bz_t^1$ ($\bar{\bz}_t^2$), followed by an online (a target) DM that outputs prediction of one-step state representation $\hat{\bx}_{t+1}^1$ ($\bar{\bx}_{t+1}^2$), given action $\ba_t$. Then, the online (target) projector maps predictions into projection space as $\hat{\by}_{t+1}^1 =g_o(\bx_{t+1}^1)$ and $\bar{\by}_{t+1}^2=g_m(\bar{\bx}_{t+1}^2)$. Introducing projection space empirically improves the overall performance.
Meanwhile, only online parameters $\theta_o$ are updated via gradient descent during the training, and the target line network parameters $\theta_m$ are updated with exponential moving average (EMA):
\begin{equation*}\label{eq:ema}
    \theta_m \leftarrow (1-\tau) \theta_m + \tau\theta_o,
\end{equation*}
where $\tau\in [0,1]$ is the EMA coefficient. 
\paragraph{Auxiliary task.} We design an auxiliary task for self-supervised representation learning along with policy learning. It contains two parts: a prediction loss $\mathcal{L}_{pre}$ and a view-consistency loss $\mathcal{L}_{con}$. Practically, we apply an prediction head $q_{pre}$ to the online projection that gives a prediction towards the future view representation $\tilde{\by}_{t+1}=g_m(f_m(\bv_{t+1}))$:  
\begin{align*}
    \mathcal{L}_{pre} = 2- 2\frac{q_{pre}(\hat{\by}_{t+1}^1)}{\|q_{pre}(\hat{\by}_{t+1}^1)\|_2}\frac{\tilde{\by}_{t+1}}{\|\tilde{\by}_{t+1}\|_2}.
\end{align*}

\noindent For view-consistency loss, we apply another predictor $q_{con}$ to the projection:
\begin{align*}\label{eq:consist}
    \mathcal{L}_{con} = 2- 2\frac{q_{con}(\hat{\by}_{t+1}^1)}{\|q_{con}(\hat{\by}_{t+1}^1)\|_2}\frac{\bar{\by}_{t+1}^2}{\|\bar{\by}_{t+1}^2\|_2}.
\end{align*}

\paragraph{Training objective.} 
Composing the RL and auxiliary objectives gives the overall training objective:
\begin{equation}\label{eq:total_loss}
    \mathcal{L}_{total}= \mathcal{L}_{rl} + \underbrace{\mathcal{L}_{pre} + \lambda\mathcal{L}_{con}}_{\text{Auxiliary Loss}}, 
\end{equation}
where  $\lambda$  steers the weight of the view-consistency loss. The auxiliary loss updates all online network parameters $\theta_o$. Note that any RL algorithm (corresponds to $\mathcal{L}_{rl}$) can be the candidate in the policy learning stage. 

\begin{figure*}
\begin{center}
\resizebox{\textwidth}{!}{\includegraphics[width=1\hsize]{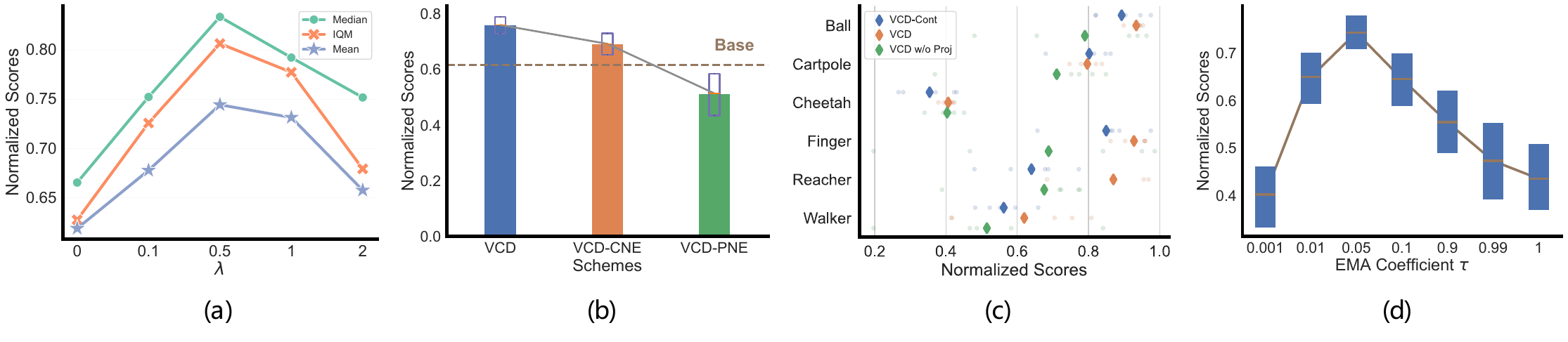}}
\vspace{-5mm}
\end{center}
\caption{Ablation studies. \textbf{(a). Aggreagte performance} of VCD with different weighting factors ($\lambda$) of view-consistency loss. \textbf{(b). Efficacy of different VCD modules}, including vanilla \textcolor{citecolor}{VCD}, VCD with a view-consistency-agnostic encoder (\textcolor{bred}{VCD-CNE}), VCD with a prediction-agnostic encoder (\textcolor{ogreen}{VCD-PNE}). \textbf{(c). Performance of VCD and its variants} including vanilla \textcolor{citecolor}{VCD}, VCD with contrastive loss (\textcolor{bred}{VCD-Cont}) and VCD without projectors (\textcolor{ogreen}{VCD w/o Proj}), where diamonds are sample-means and circles are scores of individual runs. \textbf{(d). Effect of different EMA coefficients} for target networks. All ablation experiments are conducted on six tasks on DMC-100k with 5 seed runs.}
\label{exp:ablation}
\end{figure*}

\section{Experiments}
We evaluate VCD on diverse visual controls tasks. We first outline the experiment setup, including environments and evaluation metrics in Section~\ref{exp:setup}. Then we present the training details in Section~\ref{exp:train-detail} and compare VCD with other state-of-the-arts in Section~\ref{exp:performance}. Finally, we conduct ablation studies to validate the efficacy of different VCD modules in Section~\ref{subsec:ablation}.

\subsection{Setup}
\label{exp:setup}
\paragraph{DMControl.} 

First, we evaluate the performance of VCD on six continuous control tasks in the DeepMind Control Suit (DMControl) ~\cite{tassa2018deepmind}, which is a widely-adopted benchmark for measuring the data efficiency of an RL algorithm. These tasks are of different traits and provide RL agents with image-based states. 
Following previous works~\cite{laskin2020curl}, we measure the performance of VCD at 100k and 500k environment steps during the training stage (referred to as \textbf{DMC-100k} and \textbf{DMC-500k}, respectively).\footnote{The environment step is defined as the number of environment transitions instead of learning updates (e.g., 250 learning updates with an action-repeat of $4$ corresponds to $1000$ environment steps).} 
DMC-100k investigates the data-efficiency, while DMC-500k evaluates the asymptotic long-horizon performance of an RL algorithm. The score range for each task is $[0,1000]$, which we normalize to $[0,1]$. 
\paragraph{Atari.}
Second, we test VCD on the the \textbf{Atari-100k} benchmark which consists of 26 discrete image-based controls tasks. 
The agent is allowed to play with $100$k environment steps or two hours of game-playing compared to the standard benchmark of $500$k environment steps (or 39 days of experience). The performance is measure by \textit{human-normalized score} (HNS) computed by $(s_a-s_r)/(s_h-s_r)$, where $s_a$, $s_r$ and $s_h$ denote agent score, score of random play and score of human play, respectively. 

\paragraph{Evaluation Metric.}
A fair comparison for deep RL has always been a thorny problem. Point-estimate methods like mean and median score adopted by most previous works are not suitable in the few-run cases and are not robust to outliers (extremely high (or low) score given a (mis-) fortunate seed). Recent work ~\cite{agarwal2021reliable} investigates the biases in those conventional evaluation metrics. Following their recommendation, we make three attempts here, seeking for fair comparison of VCD and its predecessors: 
\begin{inparaenum}[(1)] 
\item We run each task on both environments with 10 different seeds;
\item We report the Mean and Interquartile-mean (IQM)\footnote{IQM discards the bottom and top $25\%$ of the runs and calculates the mean score of the remaining $50\%$ runs.} score with \textit{Percentile Confidence Intervals} as an aggregate metric across all tasks and seeds;    
\item We present the \textit{Performance Profile}\footnote{Performance Profile approximates the Score Distribution $X$ as the fraction of runs above a certain normalized score $\rho$ across $N$ task and $M$ seeds, i.e., $\hat{F}_{X}(\rho)=\frac{1}{M} \sum_{m=1}^{M} \frac{1}{N} \sum_{n=1}^{N} \mathds{1}\left[x_{m, n}>\rho\right]$.} with $95\%$ confidence bound to approximate the run-score distributions; 
\end{inparaenum}
All these metrics are calculated with the open-source library \href{https://github.com/google-research/rliable}{rliable}.

\subsection{Training Details}
\label{exp:train-detail}
We largely follow previous training settings to avoid the potential impact of varying network architectures or hyperparameters. Specifically, our VCD is built on top of SAC~\cite{haarnoja2018soft} on DMControl. We adopt the encoder and actor-critic networks from ~\cite{laskin2020curl} and set DM, projectors, and predictors as MLPs. The views are generated from \textit{random crop} on the image-based states. For Atari, we follow ~\cite{schwarzer2020data} for DQN~\cite{Mnih2015HumanlevelCT} hyperparameters and network architectures, where \textit{random shift} is applied for view-generation. The main difference is adding a target DM and an independent predictor for computing view-consistency loss. Besides, we set the EMA coefficient $\tau$ for the target networks as $0.05$ and $0$ on DMControl and Atari, respectively. The weight of view-consistency loss $\lambda$ is set to $0.5$ for both environments (except for \texttt{Walker\_walk} in DMControl). We will provide an open-source implementation of VCD once accepted.

\subsection{Performance Comparison }
\label{exp:performance}

\paragraph{Comparison on DMControl. }
The left two columns in Figure~\ref{fig:main_all} compare VCD with other SOTA model-based and model-free algorithms on DMControl. For data-limited regime of DMC-100k, VCD reaches an IQM of $0.80$ and Mean of $0.75$, which is $9.5\%$ and $8.7\%$ higher than the previous best result from DrQ (an IQM of $0.73$ and Mean of $0.69$). Also, the variance of VCD is much smaller than most previous methods, which again indicates the benefit of VCD in improving data-efficiency. Similarly, the performance profile of DMC-100k, which summaries the run-score distribution of each method, shows that $75\%$ runs of VCD are above $\rho=0.6$ normalized score, compared to $63\%$ of previous best method (DrQ). A fraction of $53\%$ of runs are above $0.8$ normalized score, which is $12\%$ higher than its best predecessor DrQ ($41\%$).

In the asymptotic optimal regime of DMC-500k, VCD is competitive ($IQM=0.93, Mean=0.89$) against the previous best method DrQ ($IQM=0.92,Mean=0.88$) but with lower variance, meaning that VCD would also boost and stabilize the long-term performance of RL agents. This is also reflected in the performance profile of DMC-500k. 

\paragraph{Comparison on Atari-100k}
The rightmost column of Figure~\ref{fig:main_all} compares VCD with the other six representatives of influential algorithms. For HNS, VCD reaches an IQM of $0.38$ and Mean of $0.65$, which is $13\%$ and $5,7\%$ higher than the previous best method SPR (an IQM of $0.33$ and Mean of $0.61$). As for the performance profile across all runs, VCD mainly gains improvements on those mediocre tasks for RL agents (with HNS ranging from $0.2$ to $0.6$) compared with previous best method.

\subsection{Ablation Studies}\label{subsec:ablation}
\paragraph{View-consistency loss is critical.}
In Eqn.~\ref{eq:total_loss}, we leave the weighting of the view-consistency loss as a hyperparameter $\lambda$ that steers the loss-balancing. 
From Figure~\ref{exp:ablation}{\color{bred}{(a)}}, we can see that a positive $\lambda$ introduces a certain degree of view-consistency to the learnt dynamics and helps improve the data-efficiency w.r.t. all evaluation metrics. 
Notably, the best performance is achieved with a medium $\lambda$ at around $0.5$. Such a “concavity” indicates that too small $\lambda$ brings weak supervision signals and overwhelmingly emphasizing view-consistency will diminish the effect of other learning objectives.

To investigate view-consistency loss further, we also explore a multi-step version of VCD.
It predicts state representations $K$-steps into the future given input $\bz_t$ and a sequence of actions $(\ba_t,...,\ba_{t+K-1})$. 
Table~\ref{table:multi-steps} shows that multi-step (two-step here) prediction helps when there's no view-consistency, while incorporating multi-step prediction hampers the performance of VCD. It implies that the augmentation of view-consistency already enables DM to learn sufficiently predictive representation within only a single step. Instead, too large $K$ may unbalance the multiple losses that incur performance degeneration, a similar trend followed by $\lambda=0$ as well.
\begin{table}[H]
\vspace{-0.075in}
\caption{\textbf{Ablations on multi-step VCD.} Performance evaluation for method  \textcolor{bred}{1} (Original VCD) and \textcolor{ogreen}{2} (VCD without view-consistency) with different Predition Steps on DMC-100k with $5$ random seeds.}
\label{table:multi-steps}
\centering
\begin{small}
\begin{tabular}{lccccc}
\toprule
 & Weight $\lambda$ & Pred. Steps $K$ & Median & IQM & Mean \\\midrule
\color{bred}{1} & $0$ & $1$ & $0.665$ & $0.527$ & $0.618$ \\
\color{bred}{2} &   & $2$ & $0.726$ & $\mathbf{0.736}$ & $\mathbf{0.701}$ \\
\color{bred}{3} &   & $3$ & $\mathbf{0.731}$ & $0.685$ & $0.650$ \\
\cmidrule(lr){1-6}
\color{ogreen}{4}   & $1$ &$1$ & \cellcolor{citecolor!15}$\mathbf{0.792}$ & \cellcolor{citecolor!15}$\mathbf{0.777}$ & \cellcolor{citecolor!15}$\mathbf{0.731}$ \\
\color{ogreen}{5}   &   & $2$ & $0.736$ & $0.707$ & $0.726$ \\
\color{ogreen}{6}   &   & $3$ & $0.679$ & $0.589$ & $0.588$ \\
\bottomrule
\end{tabular}
\end{small}
\vspace{-0.075in}
\end{table}

\paragraph{How does View-Consistency work?}
In the training process, the gradients from view-consistency loss $\mathcal{L}_{con}$ would simultaneously optimize the encoder and DM. A natural question arises: \textit{how does view-consistency help the representation learning}? Is it because a view-consistent \textit{DM} captures more “realistic” dynamics in the latent space? Or is it because the encoder extracts more predictive information to improve the state representation? 
To studies this, we design another two schemes with other settings of VCD unchanged.
\begin{inparaenum}[(1)]
    \item \textit{VCD-PNE} where the encoder is not updated by the prediction loss $\mathcal{L}_{pre}$.
    \item \textit{VCD-CNE} where the encoder is not updated by the view-consistency loss $\mathcal{L}_{con}$.
\end{inparaenum}

Figure~\ref{exp:ablation}{\color{bred}{(b)}} shows that both VCD and \textit{VCD-CNE} outperform the \textit{base} scheme (VCD with $\lambda=0$) w.r.t. mean scores, while \textit{VCD-PNE} incurs performance drop. We also present relative gains of these schemes in Table~\ref{table:abl-closs-effect}. The results imply that not only that view-consistency loss contributes to training a better DM (with a relative gain of $+12.1\%$), but the main gain attributes to boosting the representative power of encoder (with a relative gain of $+23.1\%$). Meanwhile, the performance drop of \textit{VCD-PNE} demonstrates that learning a precise DM is also of significance, without which adding view-consistency may slightly impoverish the performance (with a relative gain of $-16.8\%$). These findings indicate that view-consistency aids the simultaneous training of DM and encoder to achieve great performance.  

\begin{table}[h]
\caption{\textbf{Ablations on VCD-Variants.} Relative gains of mean scores of VCD and its variants on DMC-100k with $5$ random seeds.}
\begin{scriptsize}
\centering
\begin{tabular}{cccccc}
\toprule
\multirow{2}{*}{Scheme}  & \multicolumn{2}{c}{Optimized by $\mathcal{L}_{pre}$} & \multicolumn{2}{c}{Optimized by $\mathcal{L}_{con}$} & \multirow{2}{*}{Relative Gain} \\ \cline{2-5}
        & Encoder   & Dynamics  & Encoder   & Dynamics  &     \\ \midrule
\color{citecolor}VCD     &     \Checkmark      &    \Checkmark       &  \Checkmark         &   \Checkmark        &   \cellcolor{citecolor!15}$\mathbf{+23.1\%}$  \\
\color{bred}VCD-CNE &       \Checkmark    &      \Checkmark     &           &   \Checkmark        & $\mathbf{+12.1\%}$   \\
\color{ogreen}VCD-PNE &           &  \Checkmark         &  \Checkmark         &   \Checkmark        &  $-16.8\%$  \\ \hline
\color{newbrown}Base    &     \Checkmark      &     \Checkmark      &          &           &   $0$ \\
\bottomrule
\end{tabular}
\vspace{-0.075in}
\label{table:abl-closs-effect}
\end{scriptsize}
\end{table}

\paragraph{What about constrastive loss?} 
We also implement a contrastive version of VCD (\textit{VCD-Cont}), where the losses in the auxiliary task are substituted with an InfoNCE loss following~\cite{laskin2020curl,chen2020simple}. Figure~\ref{exp:ablation}{\color{bred}{(c)}} show that VCD achieve better (or at least comparable) performance than its contrastive counterpart \textit{VCD-Cont}. We attribute this to the bias of \textit{negative samples} in the contrastive loss, as investigated by those debiased approaches~\cite{chuang2020debiased}.  Meanwhile, we observe that \textit{VCD w/o Proj} significantly outperforms original VCD, i.e., learning representation in a projection space is more preferable than learning directly in the embedding space following an encoder ~\cite{grill2020bootstrap}. 

In Table~\ref{tab:pred_steps}, we also find that providing distinct predictors for the prediction and consistency branch in VCD boosts performance. 
We attribute this to the fact that the (gradient) information of different SSL tasks may inference each other when the number of predictors becomes a bottleneck.

\begin{table}[H]
\vspace{-0.075in}
\caption{\textbf{Ablations on predictors.} Performance of VCD with different number of Predictors  on DMC-100k over 5 seeds. }
\label{tab:pred_steps}
\centering
\begin{small}
\begin{tabular}{lcccc}
\toprule
  &  Num. of Predictors & Median & IQM & Mean \\\midrule
\color{bred}{1} &  $0$ & $0.663$ & $0.646$ & $0.641$ \\
\color{bred}{2} &  $1$ & $0.720$ & $0.736$ & $0.713$ \\
\color{bred}{3} &  $2$  &  \cellcolor{citecolor!15}$\textbf{0.792}$ &  \cellcolor{citecolor!15}$\textbf{0.777}$ &  \cellcolor{citecolor!15}$\textbf{0.731}$  \\
\bottomrule
\end{tabular}
\end{small}
\vspace{-0.075in}
\end{table}

\paragraph{EMA matters.}
We consider EMA an important factor for the final performance due to the two-stream network design. Figure~\ref{exp:ablation}{\color{bred}{(d)}} indicates both updating target networks too slowly or frequently hampers model learning, while a mediocre $\tau$  around $0.05$ yields better performance. This observation is in line with the observations in~\cite{grill2020bootstrap} that a moderate EMA factor is the most suitable in SSL auxiliary tasks.

\section{Conclusion}

In this paper, we investigate the problem of data inefficiency for deep RL agents from a view representation perspective. We introduce a new framework, MMDP, to characterize the interaction between the dynamics model and view representations. Based on that, we proposed View-Consistent Dynamics to accelerate the representation learning in RL, which can be built as an auxiliary task on top of any deep RL algorithms. Finally, we validate its efficacy on diverse visual control tasks with rigorous statistical metrics.

\newpage
\bibliographystyle{named}
\bibliography{ijcai22}

\end{document}